\documentclass[conference]{IEEEtran}
\usepackage{cite}
\usepackage{amsmath,amssymb,amsfonts}
\usepackage{algorithmic}
\usepackage{graphicx}
\usepackage{textcomp}
\usepackage{xcolor}
\usepackage{verbatim}
\usepackage{hyperref}
\usepackage{caption}
\usepackage[switch]{lineno}
\def\BibTeX{{\rm B\kern-.05em{\sc i\kern-.025em b}\kern-.08em
    T\kern-.1667em\lower.7ex\hbox{E}\kern-.125emX}}
\begin{document}
\title{Linear Dimensionality Reduction for Word Embeddings in Tabular Data Classification}

\author{\IEEEauthorblockN{Liam Ressel}
\IEEEauthorblockA{\textit{ETIT-KIT, Germany}}
\and
\IEEEauthorblockN{Hamza A. A. Gardi}
\IEEEauthorblockA{\textit{IIIT at ETIT-KIT, Germany}}
}

\maketitle

\begin{abstract}
The Engineers’ Salary Prediction Challenge requires classifying salary categories into three classes based on tabular data. The job description is represented as a 300-dimensional word embedding incorporated into the tabular features, drastically increasing dimensionality. Additionally, the limited number of training samples makes classification challenging. Linear dimensionality reduction of word embeddings for tabular data classification remains underexplored. This paper studies Principal Component Analysis (PCA) and Linear Discriminant Analysis (LDA). We show that PCA, with an appropriate subspace dimension, can outperform raw embeddings. LDA without regularization performs poorly due to covariance estimation errors, but applying shrinkage improves performance significantly, even with only two dimensions. We propose Partitioned-LDA, which splits embeddings into equal-sized blocks and performs LDA separately on each, thereby reducing the size of the covariance matrices. Partitioned-LDA outperforms regular LDA and, combined with shrinkage, achieves top-10 accuracy on the competition public leaderboard. This method effectively enhances word embedding performance in tabular data classification with limited training samples.
\end{abstract}


\section{Introduction}
Salary is a key factor in the job application process. In job postings, the salary is often not disclosed, which creates uncertainty for candidates when deciding whether to apply or not. The Engineers’ Salary Prediction Challenge, held by \cite{bitgrit}, addresses this issue by focusing on the engineering job market in the United States. It requires participants to classify the salary category (low, medium, or high) based on tabular data created from the job postings. The dataset contains several features; for instance, job title, job state, and job description. The evaluation metric is accuracy.

One special aspect is that the dataset (hereafter referred to as the competition dataset) contains a vectorized job description, which is not typical for tabular data. As a result, the observations are high-dimensional; the vectorized job description increases the number of features from 15 to 315. Converting a word into a vector is referred to as word embedding. Since the vectorized job description is a block of text, each word is embedded and then averaged to create a so-called sentence embedding. However, in this research paper, we refer to it as word embedding because the properties are similar.

We want to investigate how linear dimensionality reduction methods, such as principal component analysis (PCA) or linear discriminant analysis (LDA), applied to the word embeddings, affect the performance of the overall downstream classification of tabular data. Since the dimensionality of the word embeddings is significantly reduced - for example, from 300 to 2 by performing LDA - the competition dataset now contains only 17 features and may generalize better. Furthermore, we propose a new method, Partitioned-LDA, which is expected to increase the robustness of regular LDA.

\section{Related Work\label{sec:rw}}
Word embeddings are dense, real-valued, high-dimensional vectors. Their geometry aims to capture the semantic similarity of individual words. If two words are semantically similar, such as fast and quick, their representations in the vector space should have a small distance and point in the same direction. Word embeddings are typically learned using machine learning models such as Word2Vec \cite{Word2Vec} or GloVe \cite{GloVe}. An additional challenge arises because job descriptions consist of multiple words. A simple yet surprisingly effective solution \cite{All-but-the-top} is to represent the entire job description by averaging the embeddings of all the words it contains. Furthermore, word embeddings have two interesting properties, which will be discussed below.

Let $\mathcal{D} \subset \mathbb{R}^{300}$ be the set of all non-zero job descriptions of the training set in the competition dataset. An individual job description in this set is denoted by $\boldsymbol{v} \in \mathcal{D}$. We can decompose the job description into a common mean vector $\boldsymbol{\mu} = \frac{1}{|\mathcal{D}|}\sum_{\boldsymbol{v} \in \mathcal{D}}\boldsymbol{v}$ and its remainder $\boldsymbol{r}$:
\begin{equation}
    \boldsymbol{v} = \boldsymbol{\mu} + \boldsymbol{r}
\end{equation}
According to \cite{All-but-the-top}, the common mean vector $\boldsymbol{\mu}$ is non-zero and has a relatively large norm. This can be expressed by the ratio $R$ where the denominator is the average norm of a job description:
\begin{equation}
    R = \frac{\|\boldsymbol{\mu}\|}{\frac{1}{|\mathcal{D}|}\sum_{\boldsymbol{v} \in \mathcal{D}}\|\boldsymbol{v}\|} \in [0,1]
\end{equation}
In the experiments conducted by \cite{All-but-the-top}, the inspected word embeddings attained a large ratio $R\in[\frac{1}{6}, \frac{1}{2}]$. Surprisingly, the job descriptions in our training set exhibit an enormous ratio of $R = \frac{11}{12}$, i.e., they share a common mean vector whose contribution accounts for over 90\% of the norm of each job description on average.

\begin{figure}[t]
\centering
\includegraphics[width=\columnwidth]{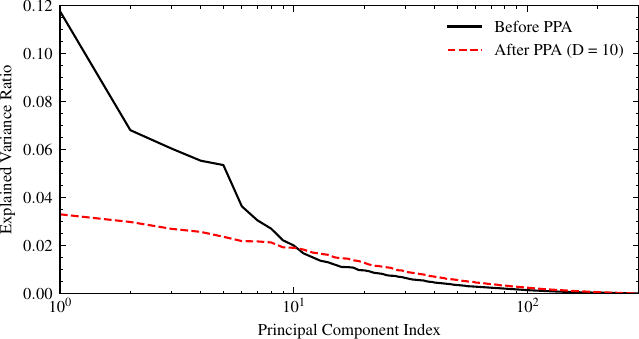}
\caption{Explained variance ratio before and after PPA}
\label{explained_variance}
\end{figure}

Furthermore, the remainder $\boldsymbol{r}$  is not isotropic, meaning that most of the energy is contained in a low-dimensional subspace \cite{All-but-the-top}. This can be analyzed using PCA. In this case, PCA is not used for dimensionality reduction but rather to demonstrate how the explained variance ratio decays exponentially, as shown in Fig.~\ref{explained_variance}. Consequently, most of the energy lies in the first few dominating principal directions, roughly $D = 10$.

The postprocessing algorithm (PPA) proposed by \cite{All-but-the-top} is based on these two observations. Since all word embeddings share the same common mean vector $\boldsymbol{\mu}$ and all remainders $\boldsymbol{r}$ have the same dominating $D$ directions, they influence the word embeddings in the same way. Therefore, eliminating $\boldsymbol{\mu}$ and the top $D$ directions from the word embeddings may improve their discriminative power. Fig.~\ref{explained_variance} illustrates the variance distribution before and after PPA, showing that the explained variance ratio no longer decays exponentially.

\section{Baseline}
Tabular data is a subcategory of heterogeneous data, where each row represents an observation and each column represents a feature. Furthermore, the features often have different data types; for example, the competition dataset includes binary, categorical, and numerical features. Another challenge is that collecting tabular data is usually expensive. As a result, low data quality is common, with issues such as missing values, erroneous observations, and class imbalance being typical for tabular data. This is also true for the competition dataset. Moreover, handling categorical features remains difficult. Lastly, the smallest possible change in a categorical or binary feature can entirely flip a prediction on tabular data. For these reasons, our model will be a gradient boosting decision tree (GBDT), as they are designed to handle the aforementioned problems and generally achieve high performance. According to \cite{Survey}, GBDT models outperform deep learning models on supervised learning tasks.

As the name Gradient Boosting Decision Tree indicates, it is an ensemble learning algorithm where many weak learners are trained sequentially. Each decision tree tries to minimize the error of the previous tree. Additionally, the gradient descent algorithm is used to control how new trees correct the errors of the existing model. State-of-the-art GBDT models include XGBoost \cite{XGBoost}, LightGBM \cite{LightGBM}, and CatBoost \cite{CatBoost}.

A popular dataset for tabular data is the adult dataset \cite{adult}. The goal is to predict whether an individual’s annual income exceeds \$50K per year based on census data. It has similar features to the competition dataset, such as occupation, native country, and education. The only significant difference is the vectorized job description. Furthermore, the competition dataset has three classes instead of two and a smaller number of training samples. Nevertheless, the adult dataset and the competition dataset are quite similar.

According to the open performance benchmark in \cite{Survey}, LightGBM outperformed all deep learning models and performed slightly better than XGBoost and CatBoost on the adult dataset. Furthermore, LightGBM speeds up the training process of conventional GBDT models by over 20 times while achieving almost the same accuracy \cite{LightGBM}. Because of this, we use the histogram-based gradient boosting classification tree (HGBClassifier) from scikit-learn \cite{scikit}, since its implementation is inspired by LightGBM. It handles categorical features automatically when explicitly specified. Furthermore, it has built-in support for missing values. In general, GBDT models also perform automatic feature selection to a sufficient extent, and feature scaling is not needed because they are based on decision trees. Therefore, data preprocessing involves only identifying and listing the categorical features.

This concludes the baseline, i.e., the HGBClassifier, where the vectorized job descriptions are fed into the model without any postprocessing. Additionally, default parameters with categorical feature support are used. This simple setting, without any cumbersome data preprocessing or cleaning, achieved an accuracy of 74.00\% on the public leaderboard.

\begin{figure}[t]
\centering
\includegraphics[width=\columnwidth]{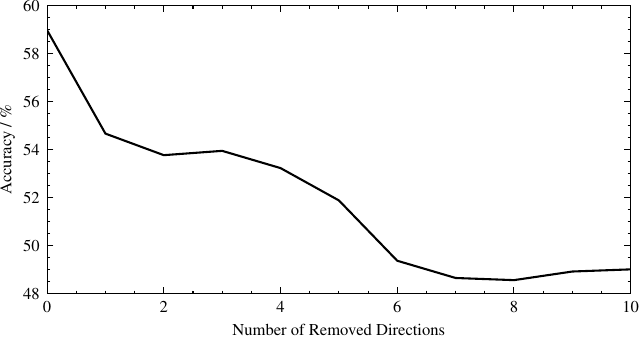}
\caption{Relationship between accuracy and number of removed directions}
\label{PPA}
\end{figure}

\section{Methods\label{sec:methods}}
We investigate methods applied exclusively to the job descriptions. Logistic regression is used for evaluation because it better reflects positive changes due to postprocessing, given that it is a linear classifier. Additionally, we use the training set from the competition dataset with stratified 5-fold cross-validation to assess performance. Without postprocessing, logistic regression achieves an accuracy of 59.16\%.

\begin{figure}[t]
\centering
\includegraphics[width=\columnwidth]{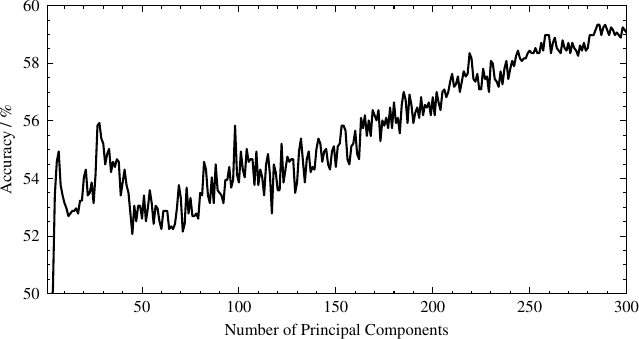}
\caption{Relationship between accuracy and number of principal components}
\label{PCA}
\end{figure}

\subsection{Postprocessing Algorithm}
As mentioned in Section~\ref{sec:rw}, PPA removes the common mean vector and eliminates the dominating directions. This is done by projecting the top $D$ principal directions $\boldsymbol{u}_1,\boldsymbol{u}_2,\ldots,\boldsymbol{u}_D$  away from the centered job description $\boldsymbol{r} = \boldsymbol{v} - \boldsymbol{\mu}$:
\begin{equation}
    \tilde{\boldsymbol{r}} = \boldsymbol{r} - \sum_{i=1}^{D} (\boldsymbol{u}_i^\mathrm{T}\boldsymbol{r})\boldsymbol{u}_i
\end{equation}
This could increase the discriminative power, as the job descriptions no longer share a large common mean vector $\boldsymbol{\mu}$ and are not influenced by the top $D$ directions in the same way (see Fig.~\ref{explained_variance}). According to \cite{All-but-the-top}, PPA is empirically validated on lexical and sentence-level tasks across multiple datasets. However, it remains unclear whether these findings also generalize to a downstream classification task, where the vectorized job descriptions are used as features. 

Fig.~\ref{PPA} displays the result of PPA. It shows that removing the top $D$ directions significantly decreases accuracy, indicating that the discriminative information resides in those removed dominant directions.

\subsection{Principal Component Analysis}
Since it has turned out that the top $D$ principal directions carry significant discriminative information, it may be helpful to keep them and discard those with small variance to reduce noise. But first, let us recap what PCA actually does.

PCA linearly projects the data onto a lower-dimensional subspace, maximizing the variance of the projected data. Let $N$ be the number of observations in our training set. First, we have to estimate the data covariance matrix:
\begin{equation}
    \hat{\boldsymbol{\mathrm{C}}} = \frac{1}{N}\sum_{i=1}^{N}\boldsymbol{r}_i\boldsymbol{r}_i^\mathrm{T} 
\end{equation}
Next, we compute the eigenvalues $\lambda_i$, which are real and non-negative, and arrange them in descending order. These correspond to the variance explained by each principal component. Finally, we transform our data to obtain the principal components. To do this, we also need to calculate the eigenvectors (principal directions) $\boldsymbol{u}_i$. Suppose we want to project our job descriptions onto the subspace spanned by the top $D$ directions. Then, we perform the following calculation:
\begin{equation}
    \tilde{\boldsymbol{r}} = \sum_{i=1}^{D}(\boldsymbol{u}_i^\mathrm{T}\boldsymbol{r})\boldsymbol{e}_i
\end{equation}
In summary, PCA projects data onto the top $D$ principal directions, while PPA removes them.

As we can see in Fig.~\ref{PCA}, the accuracy increases significantly in the first ten principal components, indicating that they contain a lot of discriminative information. It then increases slowly but unfortunately does not surpass the accuracy achieved by the unprocessed job descriptions.

\begin{figure}[t]
\centering
\includegraphics[width=\columnwidth]{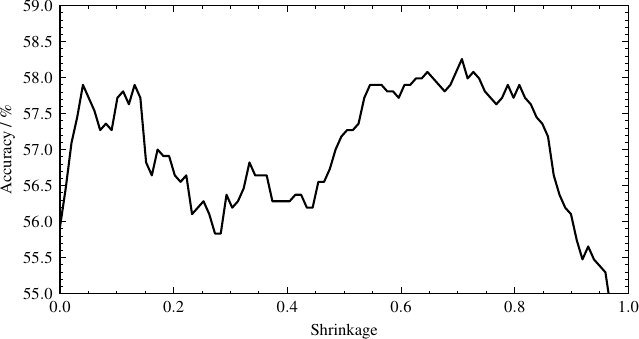}
\caption{Relationship between accuracy and shrinkage}
\label{shrinkage}
\end{figure}

\subsection{Linear Discriminant Analysis}
PPA and PCA lack a crucial aspect in the context of downstream classification tasks: they do not incorporate class label information. In contrast, LDA leverages class labels to maximize the separation between different classes. Let $K$ be the number of classes. LDA projects the data onto a subspace of $K - 1$ classes:
\begin{equation}
        \tilde{\boldsymbol{v}} = \sum_{i=1}^{K-1}(\boldsymbol{w}_i^\mathrm{T}\boldsymbol{v})\boldsymbol{e}_i = \boldsymbol{\mathrm{W}}^\mathrm{T}\boldsymbol{v}
\end{equation}
It is assumed that the dimensionality of $\boldsymbol{v}$ is greater than the number of classes, $K$.

We have three classes (low, medium, and high). Let us partition our training set into these classes: $\mathcal{D} = \mathcal{C}_\mathrm{l} \cup \mathcal{C}_\mathrm{m} \cup \mathcal{C}_\mathrm{h}$.
The next step is to calculate the class means, which are denoted by $\boldsymbol{\mu}_\mathrm{l}$, $\boldsymbol{\mu}_\mathrm{m}$, and $\boldsymbol{\mu}_\mathrm{h}$, respectively. After that, we need to estimate two covariance matrices. The within-class covariance matrix measures the variance within each class relative to its respective class mean.
\begin{equation}
\begin{split}
    \hat{\boldsymbol{\mathrm{C}}}_\mathrm{W} = & \sum_{i \in \mathcal{C}_\mathrm{l}} 
    (\boldsymbol{v}_i - \boldsymbol{\mu}_\mathrm{l})(\boldsymbol{v}_i - \boldsymbol{\mu}_\mathrm{l})^\mathrm{T} + \\
    & \sum_{i \in \mathcal{C}_\mathrm{m}} 
    (\boldsymbol{v}_i - \boldsymbol{\mu}_\mathrm{m})(\boldsymbol{v}_i - \boldsymbol{\mu}_\mathrm{m})^\mathrm{T} + \\
    & \sum_{i \in \mathcal{C}_\mathrm{h}} 
    (\boldsymbol{v}_i - \boldsymbol{\mu}_\mathrm{h})(\boldsymbol{v}_i - \boldsymbol{\mu}_\mathrm{h})^\mathrm{T}
\end{split}
\end{equation}
The between-class covariance matrix measures how far the class means are from the overall mean:
\begin{equation}
\begin{split}
    \hat{\boldsymbol{\mathrm{C}}}_\mathrm{B} = & |\mathcal{\mathcal{C}_\mathrm{l}}|\sum_{i \in \mathcal{C}_\mathrm{l}} 
    (\boldsymbol{\mu}_\mathrm{l} - \boldsymbol{\mu})(\boldsymbol{\mu}_\mathrm{l} - \boldsymbol{\mu})^\mathrm{T} + \\
    & |\mathcal{\mathcal{C}_\mathrm{m}}|\sum_{i \in \mathcal{C}_\mathrm{m}} 
    (\boldsymbol{\mu}_\mathrm{m} - \boldsymbol{\mu})(\boldsymbol{\mu}_\mathrm{m} - \boldsymbol{\mu})^\mathrm{T} + \\
    & |\mathcal{\mathcal{C}_\mathrm{h}}|\sum_{i \in \mathcal{C}_\mathrm{h}} 
    (\boldsymbol{\mu}_\mathrm{h} - \boldsymbol{\mu})(\boldsymbol{\mu}_\mathrm{h} - \boldsymbol{\mu})^\mathrm{T}
\end{split}
\end{equation}
Lastly, by maximizing the so-called Rayleigh quotient
\begin{equation}
    \boldsymbol{\mathrm{W}}_\mathrm{opt} = \arg\max_{\boldsymbol{\mathrm{W}}}\frac{\det(\boldsymbol{\mathrm{W}}^\mathrm{T}
    \hat{\boldsymbol{\mathrm{C}}}_\mathrm{B}\boldsymbol{\mathrm{W}})}{\det(\boldsymbol{\mathrm{W}}^\mathrm{T}
    \hat{\boldsymbol{\mathrm{C}}}_\mathrm{W}\boldsymbol{\mathrm{W}})}
\end{equation}
we obtain the optimal weight matrix $\boldsymbol{\mathrm{W}}_\mathrm{opt}$, which linearly projects the data such that the separation between class means is maximized while the variance within each class is minimized. For mathematical reasons, the subspace dimensionality should be smaller than the number of classes $K$. Projecting onto larger subspaces does not affect the outcome of the optimization problem.

We now fit LDA from scikit-learn \cite{scikit} to the training set and transform the training and validation sets using the learned weight matrix $\boldsymbol{\mathrm{W}}_\mathrm{opt}$. Next, we replace the job descriptions with the two transformed components and evaluate the accuracy, which drops to 55.92\% (compared to 59.16\% achieved with the raw job descriptions). LDA risks overfitting, resulting in high training accuracy but reduced validation performance. The main reason for this is that the estimation of the covariance matrices is numerically ill-conditioned due to their high dimensionality (\(300 \times 300\)) and the low number of training samples. To obtain the optimal weight matrix $\boldsymbol{\mathrm{W}}_\mathrm{opt}$, we need to invert a covariance matrix, which dramatically amplifies estimation errors \cite{Cov}. In the following, two solutions to this problem are presented.

\begin{figure}[t]
\centering
\includegraphics[width=\columnwidth]{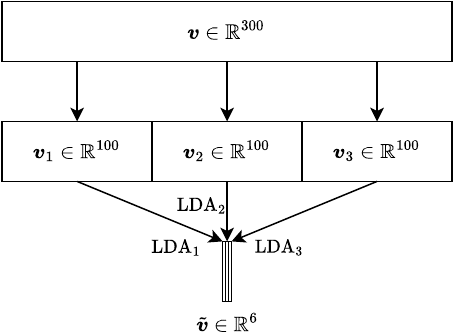}
\caption{Partitioned-LDA example}
\label{partition}
\end{figure}

\subsubsection{Shrinkage}
The estimation proposed by \cite{Honey} uses a transformation called shrinkage on the sample covariance matrix. This technique adjusts extreme values towards more central values, thereby reducing amplified estimation errors. There is one hyperparameter, namely the shrinkage constant $\delta\in[0,1]$, which influences the estimation in the following way:
\begin{equation}
    \hat{\boldsymbol{\mathrm{S}}} = \delta \hat{\boldsymbol{\mathrm{F}}} + (1 - \delta) \hat{\boldsymbol{\mathrm{C}}}
\end{equation}
Hence, the new estimator $\hat{\boldsymbol{\mathrm{S}}}$ is a linear combination of a structured estimator $\hat{\boldsymbol{\mathrm{F}}}$ and the sample covariance matrix $\hat{\boldsymbol{\mathrm{C}}}$. An example of a structured estimator is an identity matrix scaled by the average of the diagonal elements of the sample covariance matrix. The subsequent step involves the variation of the shrinkage parameter, which is also available in the scikit-learn framework. As shown in Fig.~\ref{shrinkage}, the accuracy increased by 2\% around $\delta = 0.7$, but it still falls short by 1\% compared to the raw job descriptions. Nevertheless, the result is satisfactory, as the dimensionality has been reduced by 298 dimensions.
\subsubsection{Partitioned-LDA}
\begin{figure}[t]
\centering
\includegraphics[width=\columnwidth]{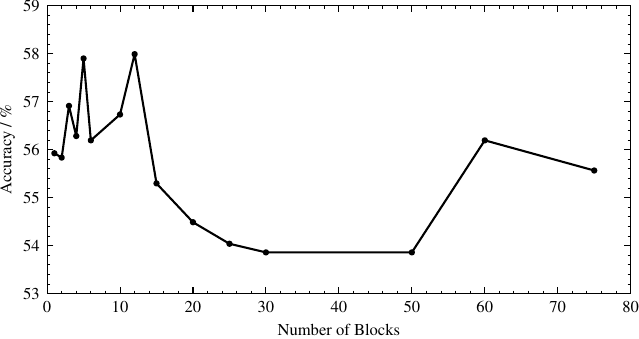}
\caption{Relationship between accuracy and number of blocks}
\label{partition_performance}
\end{figure}

We propose a new method, Partitioned-LDA, which splits the job description $\boldsymbol{v} \in \mathbb{R}^{300}$ into $N_b$ equal-sized, non-overlapping blocks, denoted as $\boldsymbol{v}_1, \boldsymbol{v}_2, \ldots, \boldsymbol{v}_{N_b} \in \mathbb{R}^s$. We only consider $N_b$ values where the job description can be fully partitioned without remainder, ensuring that $300 \mod N_b = 0$. For each individual block of size $s = \frac{300}{N_b} > 3$, an LDA is fitted to the corresponding data. An example of this method with $N_b = 3$ blocks, each of size $s = 100$, is depicted in Fig.~\ref{partition}. Consequently, this approach reduces the dimensionality from 300 to $2N_b$. The advantage of this method is that it reduces the size of the covariance matrices to $s \times s$, thereby improving the ratio of the number of training samples to the number of features. Hence, the estimation is more robust. However, this approach captures only local patterns, potentially losing some global interactions between features. Therefore, selecting an appropriate number of blocks is crucial. As shown in Fig.~\ref{partition_performance}, the optimal configuration in this case is $N_b = 12$, corresponding to a block size of $s = 25$.

\begin{figure*}[t]
\centering
\includegraphics[width=\textwidth]{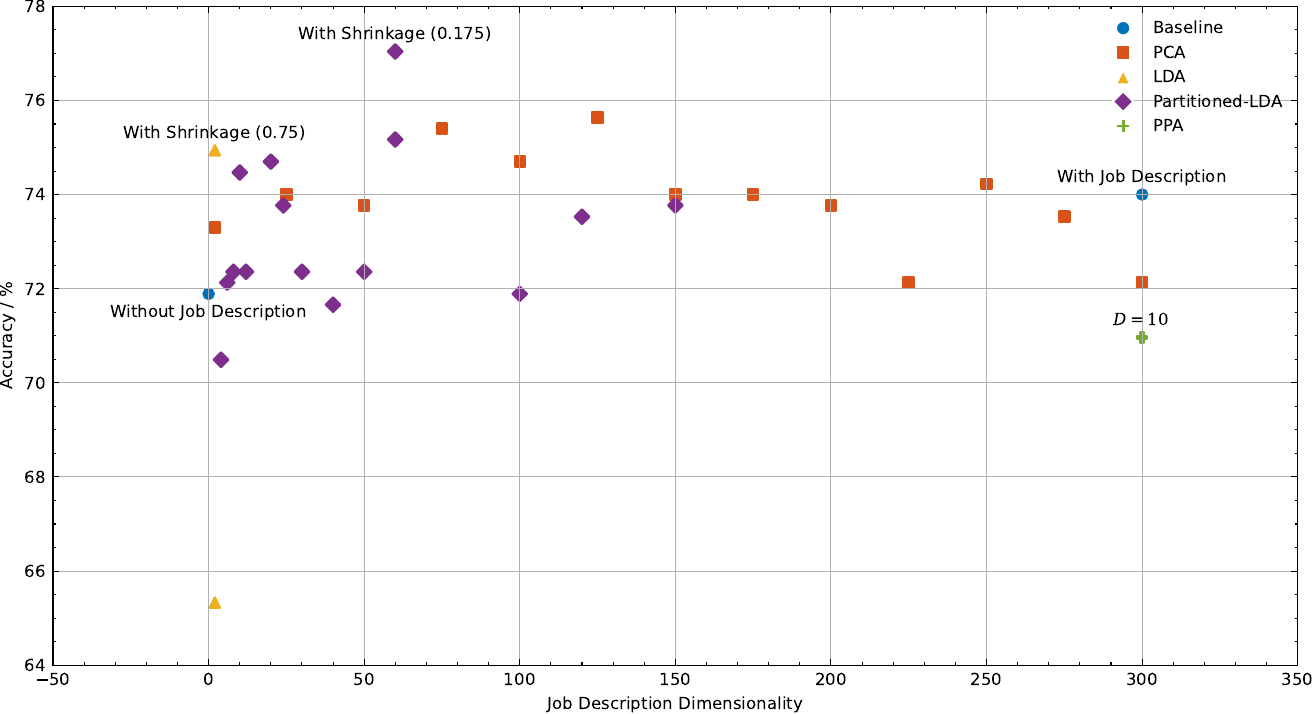}
\caption{Comparison of method performances on public leaderboard}
\label{experiment}
\end{figure*}

\section{Experiment\label{sec:experiment}}
In Section~\ref{sec:methods}, no method could outperform the raw job descriptions since word embeddings are already highly informative. Dimensionality reduction will inevitably lead to some information loss. However, if we can keep this loss low while drastically reducing dimensionality, it could be advantageous. This is especially important because they only achieve an accuracy of 59.16\% without postprocessing, which is quite poor. Moreover, a too high dimensionality may distract the classifier's attention.

In this section, we evaluate the different methods on the public leaderboard. The raw job descriptions are replaced with the outputs of the respective methods. Additionally, the HGBClassifier is employed for classification, utilizing its categorical feature support and default parameters.

We can see in Fig.~\ref{experiment} that, as expected, PPA decreases the accuracy compared to the baseline since the discriminative principal directions were removed. PCA achieves a slightly better accuracy than the baseline but performs worse at very low or very high subspace dimensions. LDA performs very poorly, decreasing the accuracy by almost 9\%. It clearly overfitted the training set and learned false patterns. Nevertheless, with an appropriate shrinkage value, chosen based on Fig.~\ref{shrinkage}, it improved the baseline accuracy by 1\% while reducing the dimensionality to only two dimensions. Partitioned-LDA initially performs better than LDA, but its average performance is slightly below the baseline. However, with the optimal number of blocks, here \( N_b = 30 \), which corresponds to a block size of \( s = 10 \), it surpassed LDA with shrinkage. Finally, by adding shrinkage to Partitioned-LDA, the accuracy further increased to 77.04\% (placing it in the top 10 on the public leaderboard).

\section{Conclusion}
Incorporating word embeddings into tabular data classification has not been properly explored. We demonstrated that linear dimensionality reduction can enhance performance if the right parameters are chosen. In PCA, the projected subspace should not be excessively low-dimensional or high-dimensional. LDA, without any regularization, performs poorly because the estimation of high-dimensional covariance matrices requires a large number of training samples. However, by adding shrinkage, LDA can outperform raw word embeddings even with only two dimensions. By splitting the word embeddings into equal-sized blocks, our proposed method, Partitioned-LDA, addresses this issue, as the covariance matrices are reduced to the size of a block. Nevertheless, there is a trade-off between the size of the covariance matrices and the global context, as LDA is performed multiple times locally. With the right block size, Partitioned-LDA can significantly enhance performance, and with additional shrinkage, it even ranked in the top 10 on the public leaderboard. Nevertheless, when word embeddings are not considered in the context of tabular data, as described in Section~\ref{sec:methods}, they can achieve optimal performance in their raw form due to their inherently high informativeness. However, in our case, their quality was suboptimal, making dimensionality reduction essential to redirect the classifier’s focus and thereby enhance overall performance. Future work could explore nonlinear dimensionality reduction methods or approaches based on manifold learning to capture more expressive features. Additionally, deep learning techniques, such as autoencoders, could be utilized for dimensionality reduction.


\begin{thebibliography}{00}

\bibitem{bitgrit} bitgrit, Engineers' Salary Prediction Challenge. [Online]. Available: https://bitgrit.net

\bibitem{Word2Vec} T. Mikolov, K. Chen, G. Corrado, and J. Dean, ``Efficient estimation of word representations in vector space,'' arXiv preprint arXiv:1301.3781, 2013.

\bibitem{GloVe} J. Pennington, R. Socher, and C. D. Manning, ``Glove: Global vectors for word representation,'' in *Proc. 2014 Conf. Empirical Methods in Natural Language Processing (EMNLP)*, 2014.

\bibitem{All-but-the-top} J. Mu, S. Bhat, and P. Viswanath, ``All-but-the-top: Simple and effective postprocessing for word representations,'' arXiv preprint arXiv:1702.01417, 2017.

\bibitem{Survey} V. Borisov et al., ``Deep neural networks and tabular data: A survey,'' *IEEE Trans. Neural Netw. Learn. Syst.*, in press.

\bibitem{XGBoost} T. Chen and C. Guestrin, ``XGBoost: A scalable tree boosting system,'' in *Proc. 22nd ACM SIGKDD Int. Conf. Knowl. Discov. Data Min.*, 2016.

\bibitem{LightGBM} G. Ke et al., ``LightGBM: A highly efficient gradient boosting decision tree,'' in *Adv. Neural Inf. Process. Syst.*, vol. 30, 2017.

\bibitem{CatBoost} L. Prokhorenkova et al., ``CatBoost: Unbiased boosting with categorical features,'' in *Adv. Neural Inf. Process. Syst.*, vol. 31, 2018.

\bibitem{adult} UCI Machine Learning Repository, Adult dataset. [Online]. Available: https://archive.ics.uci.edu/dataset/2/adult

\bibitem{scikit} F. Pedregosa et al., ``Scikit-learn: Machine learning in Python,'' *J. Mach. Learn. Res.*, vol. 12, pp. 2825--2830, 2011.

\bibitem{Cov} O. Ledoit and M. Wolf, ``A well-conditioned estimator for large-dimensional covariance matrices,'' *J. Multivariate Anal.*, vol. 88, no. 2, pp. 365--411, 2004.

\bibitem{Honey} O. Ledoit and M. Wolf, ``Honey, I shrunk the sample covariance matrix,'' UPF Economics and Business Working Paper, no. 691, 2003.

\end{thebibliography}
\end{document}